\documentclass{article}
\usepackage{0_other/ijcai23}

\usepackage{graphicx}
\usepackage{subfigure}
\usepackage{booktabs}
\usepackage{multirow}
\usepackage{pifont}
\usepackage{amsmath, amssymb, amsfonts}

\usepackage{times}
\usepackage{soul}
\usepackage{url}
\usepackage[hidelinks]{hyperref}
\usepackage[utf8]{inputenc}
\usepackage[small]{caption}
\usepackage{amsthm}
\usepackage{algorithm}
\usepackage{algorithmic}
\usepackage[switch]{lineno}

\title{Towards Accurate Acne Detection via Decoupled Sequential Detection Head}

\author{
	Xin Wei$^1$\and
	Lei Zhang$^1$\and
	Jianwei Zhang$^1$\and
	Junyou Wang$^1$\and
	Wenjie Liu$^1$\and
	Jiaqi Li$^2$\And
	Xian Jiang$^2$
	\affiliations
	$^1$College of Computer Science, Sichuan University, Chengdu 610065, China\\
	$^2$Department of Dermatology, West China Hospital, Sichuan University, Chengdu	610041, China\\
	\emails
	weixin@stu.scu.edu.cn,
	leizhang@scu.edu.cn,
	\{zhangjianwei, wangjunyou, liuwj\}@stu.scu.edu.cn,
	lijiaqicd@gmail.com,
	youradrian@outlook.com
}

\begin{document}
	
\maketitle

\begin{abstract}
Accurate acne detection plays a crucial role in acquiring precise diagnosis and conducting proper therapy. However, the ambiguous boundaries and arbitrary dimensions of acne lesions severely limit the performance of existing methods. In this paper, we address these challenges via a novel Decoupled Sequential Detection Head (DSDH), which can be easily adopted by mainstream two-stage detectors. DSDH brings two simple but effective improvements to acne detection. Firstly, the offset and scaling tasks are explicitly introduced, and their incompatibility is settled by our task-decouple mechanism, which improves the capability of predicting the location and size of acne lesions. Second, we propose the task-sequence mechanism, and execute offset and scaling sequentially to gain a more comprehensive insight into the dimensions of acne lesions. In addition, we build a high-quality acne detection dataset named ACNE-DET to verify the effectiveness of DSDH. Experiments on ACNE-DET and the public benchmark ACNE04 show that our method outperforms the state-of-the-art methods by significant margins. Our code and dataset are publicly available at (temporarily anonymous).
\end{abstract}

\begin{figure}[!t]
	\centering
	\includegraphics[width=0.96\columnwidth]{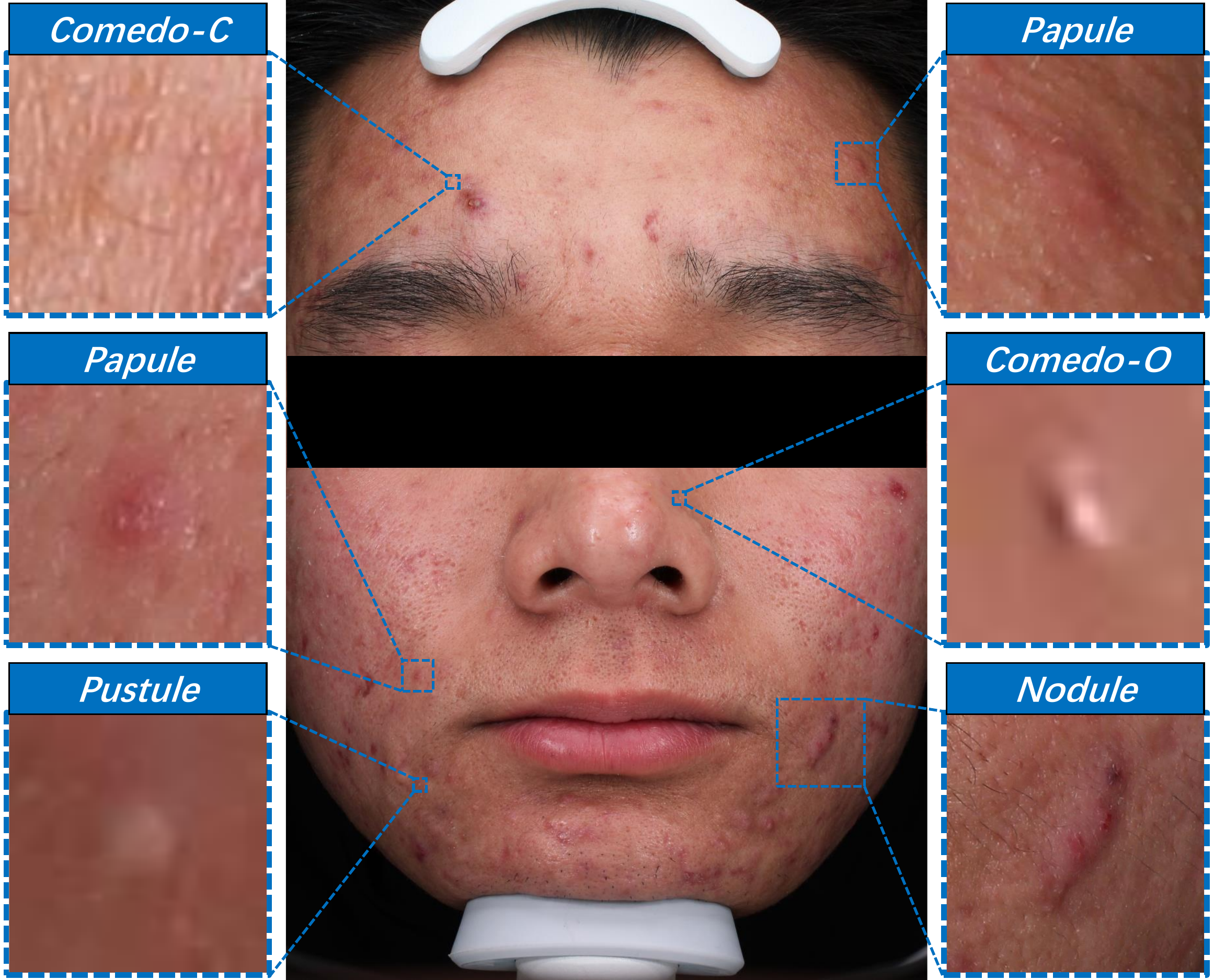}
	\caption{An example of the facial skin images from \emph{ACNE-DET}. Some acne lesion instances of the example are magnified and shown on the left and right for better clarity.}
	\label{fig_444}
\end{figure}

\section{Introduction}
Acne vulgaris, commonly known as acne, is a chronic skin disease characterized by the inflammatory process of hair follicles and sebaceous glands \cite{zaenglein2018acne}. Epidemiology shows that about 85\% of adolescents suffer from acne \cite{heng2020systematic}, which possibly persists through adulthood \cite{han2016epidemiology}. Early intervention is essential for acne patients. Otherwise, both their physical condition and psychological state can be damaged \cite{dreno2018female,hazarika2016psychosocial}. Understanding the status of the facial skin lesions is the basis of evaluating acne severity and conducting further treatment. Nevertheless, since acne patients may need periodic diagnoses, the massive consultation demands are far from being met by dermatologists \cite{clark2018acne}. Therefore, automatic and accurate acne detection algorithms are urgently needed.

Traditional acne detection mainly relies on image processing algorithms, which are built on handcrafted features, e.g., color model \cite{kittigul2016automatic}, texture-based features \cite{alamdari2016detection} and composited features~\cite{maroni2017automated}. But these methods are limited by their weak detection performance and poor generalization capability.

Convolutional neural network (CNN) based methods have achieved significant progress in detecting biomedical objects, such as glomerular \cite{nguyen2021circle}, retinopathy \cite{wang2017zoom} and universal lesion detection \cite{yan2020learning}. Despite the extensive researches in medical image detection, little attention is paid to acne detection. As a result, only preliminary studies have been made. Wu \emph{et~al}.~\shortcite{wu2019joint} construct an acne dataset named \emph{ACNE04}, which consists of images with detection annotations of a single lesion category, then conduct global acne grading and counting via label distribution learning. Wen \emph{et~al}.~\shortcite{wen2022acne} and Huynh \emph{et~al}.~\shortcite{huynh2022automatic} introduce several generic detection models into acne detection and produce baseline results. Min \emph{et~al}.~\shortcite{min2021acnet} propose an acne detection network called ACNet consisting of three components, which focus on the issues of imbalanced illumination, lesion variation and dense arrangement, respectively. 

However, none of the existing methods effectively address the following challenges: (\emph{i}) Ambiguous boundaries. There is no clear line of demarcation between the acne lesions and the surrounding normal skin, therefore, even dermatologists sometimes fail to give an accurate answer on the location and size of an acne lesion. (\emph{ii}) Arbitrary dimensions. A large acne lesion may occupy an entire patch of skin, while a small acne lesion can be hard to see with naked eyes, which implies the dimensions of the acne lesions vary widely. (\emph{iii}) Lack of datasets. Publicly available acne datasets, especially high-quality ones, are extremely scarce due to the complexity of whole-face annotation, thus severely impeding the development of acne detection. Some acne lesion instances are presented in Figure~\ref{fig_444} for illustration.

Based on the above discussions, the limited performance of existing acne detection methods motivates us to optimize the architectures of generic detection models and construct a new dataset for acne detection. The main contributions of this paper are summarized as follows:

\begin{itemize}
	\item We explicitly introduce the offset task and the scaling task, and decouple them by two standalone branches to settle their incompatibility. The task-decouple mechanism effectively improves the capability of the detection head to predict the location and size of the acne lesions.
	\item We split the bounding box refinement task into two steps, and execute the offset task and the scaling task sequentially to facilitate scaling. The task-sequence mechanism enables the detection head to gain a more comprehensive insight into the dimensions of the acne lesions. 
	\item We build a high-quality acne detection dataset named \emph{ACNE-DET} to verify the effectiveness of our methods. \emph{ACNE-DET} surpasses the previous acne dataset for its high-resolution facial skin images, superior annotation accuracy and fine-grained skin lesion categories. 
\end{itemize}

By combining the task-decouple and task-sequence mechanisms, we propose a novel Decoupled Sequential Detection Head (DSDH), which can be easily adopted by mainstream two-stage detectors. Comprehensive experiments on both \emph{ACNE-DET} and \emph{ACNE04} show that our method outperforms the state-of-the-art methods by significant margins.

\begin{figure*}[!t]
	\centering
	\subfigure[Single Detection Head]{
		\includegraphics[width=0.32\textwidth]{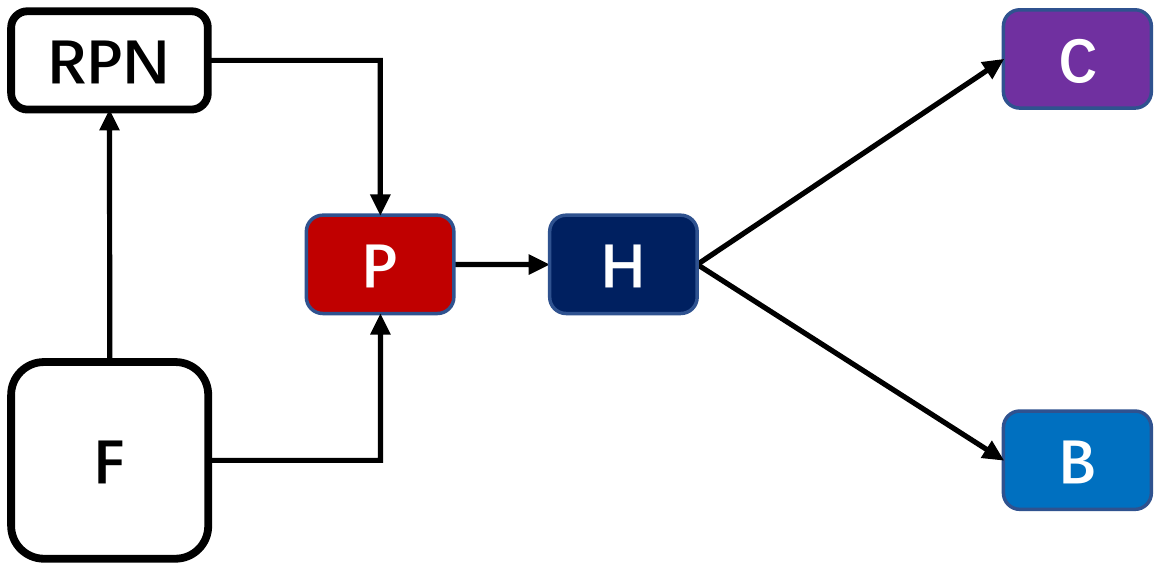}
		\label{fig_111_111}
	}
	\subfigure[Double Detection Head]{
		\includegraphics[width=0.32\textwidth]{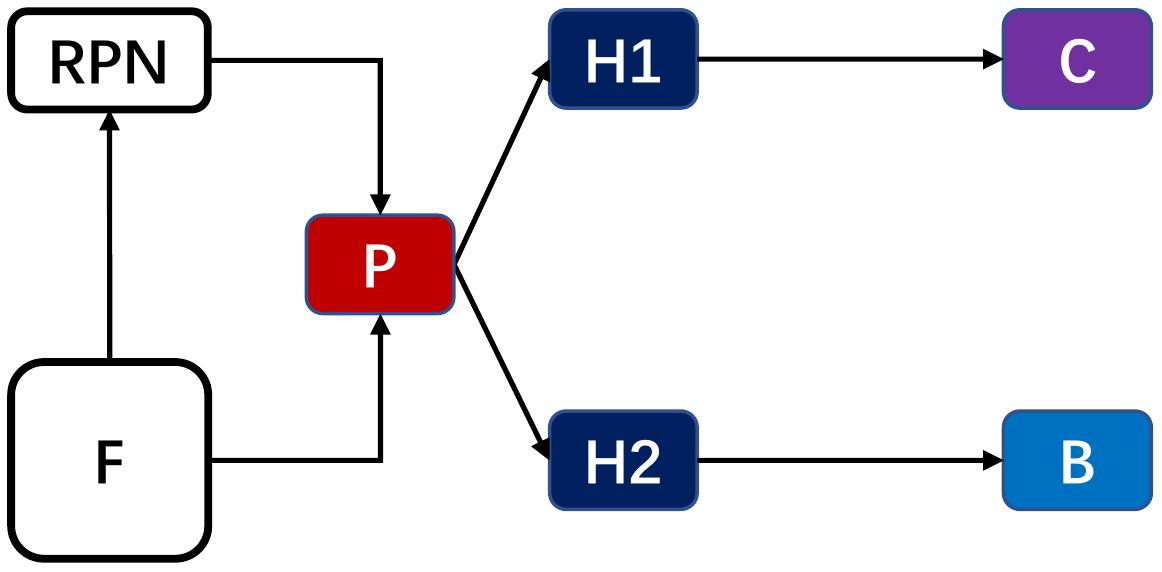}
		\label{fig_111_222}
	}
	\subfigure[Decoupled Detection Head\dag]{
		\includegraphics[width=0.32\textwidth]{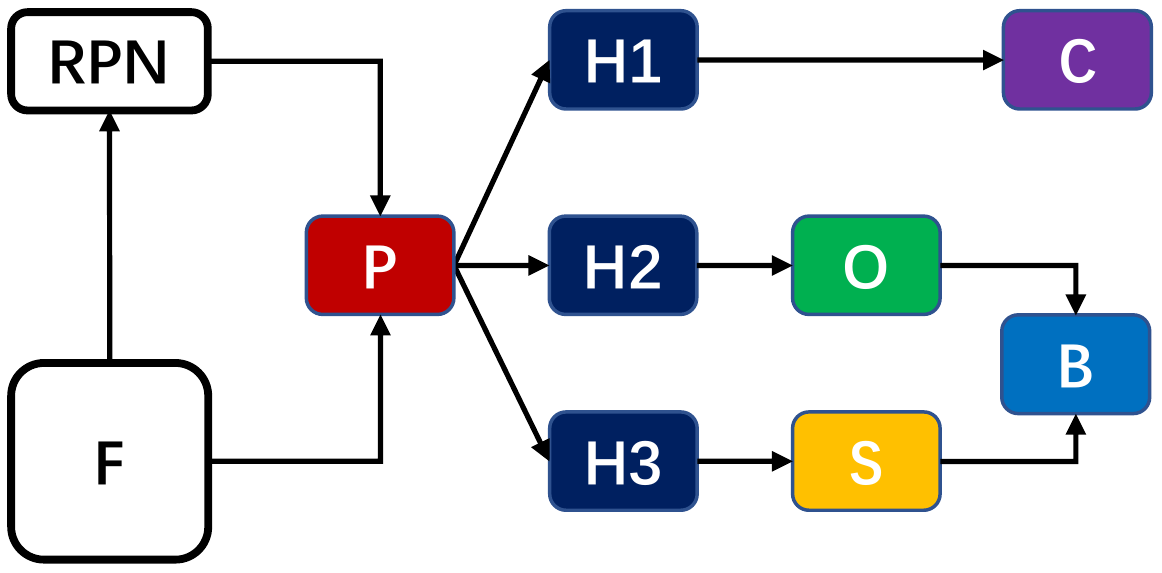}
		\label{fig_111_333}
	}
	
	\subfigure[Sequential Detection Head\dag]{
		\includegraphics[width=0.32\textwidth]{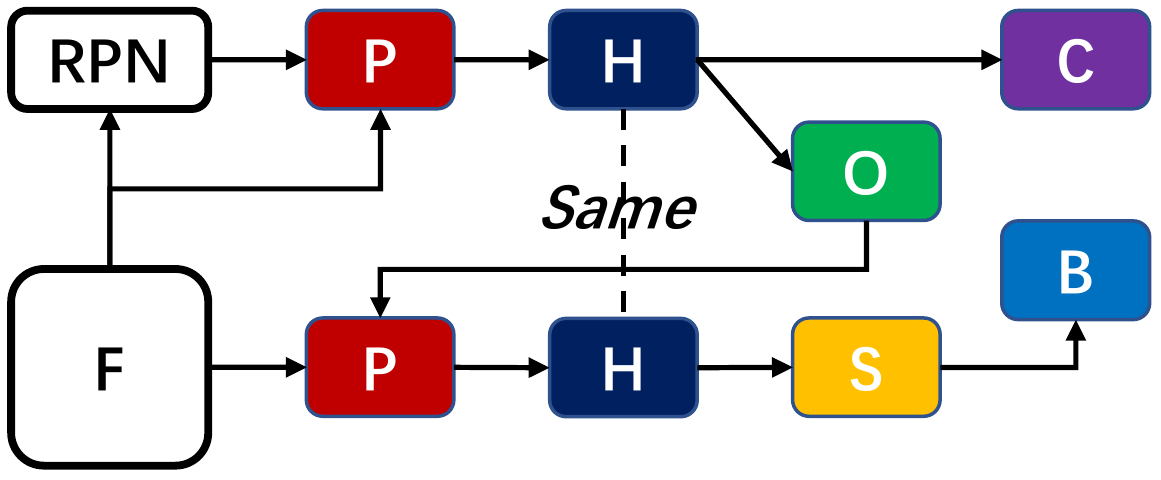}
		\label{fig_111_444}
	}
	\subfigure[Decoupled Sequential Detection Head\dag]{
		\includegraphics[width=0.32\textwidth]{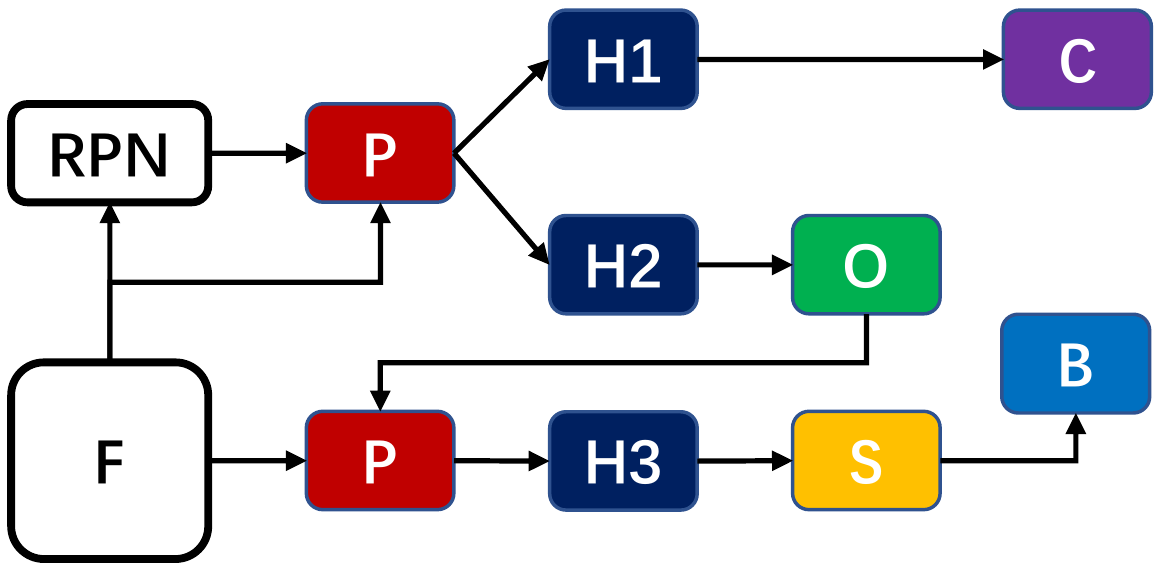}
		\label{fig_111_555}
	}
	\subfigure[Decoupled Sequential Detection Head\dag\quad (cascade version)]{
		\includegraphics[width=0.32\textwidth]{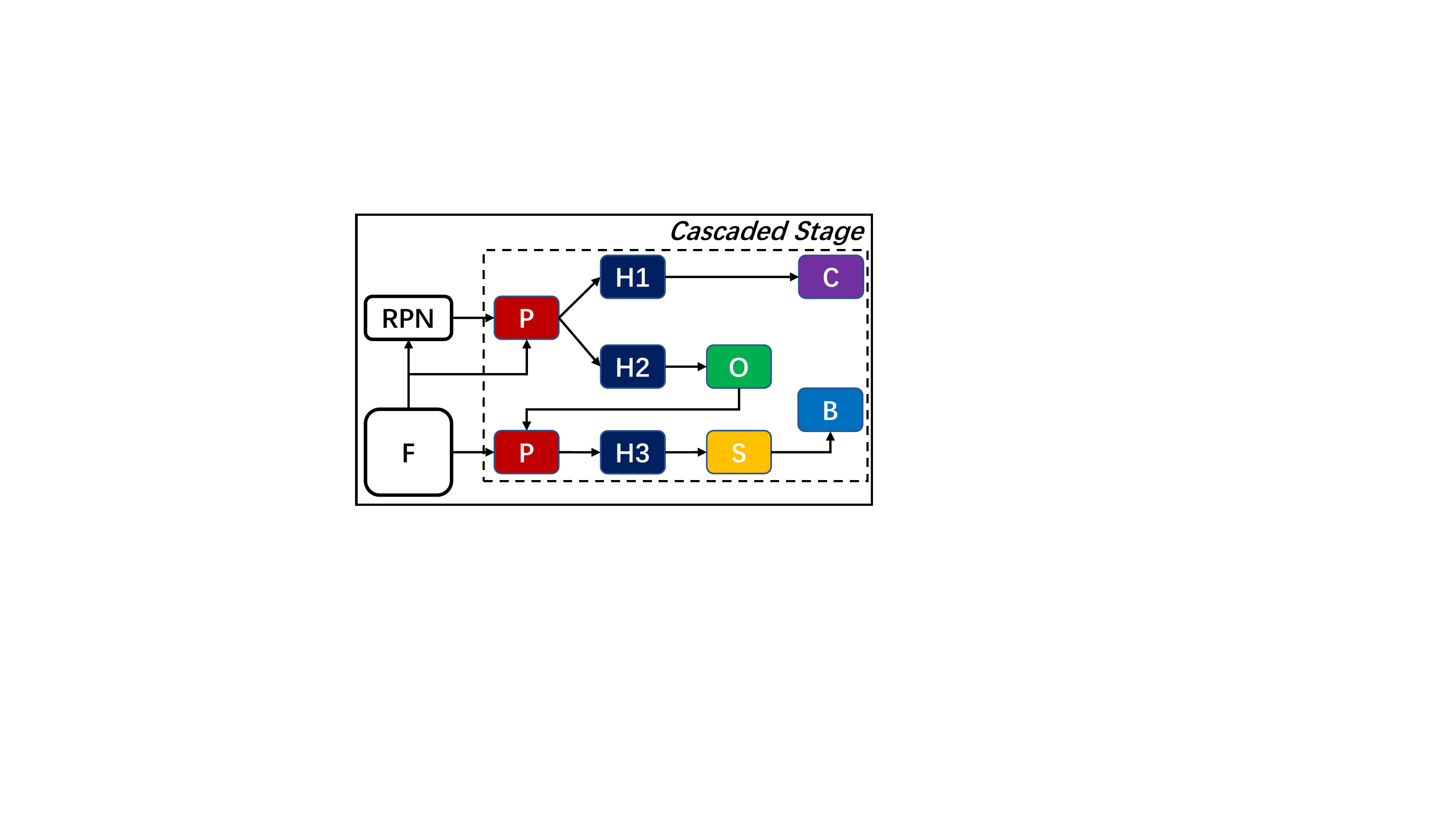}
		\label{fig_111_666}
	}
	\caption{Comparisons of different detection head architectures. “F” for feature maps extracted by backbone. “RPN” for a region proposal network. “P” for a pooling operator, e.g., RoI-pooling or RoI-align. “H” for a network branch of detection head. “C” for classification results. “B” for bounding box results. “O” for offset results. “S” for scaling results. \dag~indicates our methods.}
	\label{fig_111}
\end{figure*}

\section{Related Works}

\subsection{Object Detection}
Recent object detection algorithms are mainly based on CNN, and are divided into two categories according to their network architectures.

\paragraph{One-stage Methods.} One-stage methods formulate object detection as an end-to-end task. Thus, the classification and bounding box results are obtained in a single network. YOLO \cite{redmon2016you} introduces a unified network to object detection for the first time. SSD \cite{liu2016ssd} optimizes the object matching quality by generating anchor boxes of different sizes and aspect ratios. RetinaNet \cite{lin2017focal} constructs focal loss, which alleviates the foreground-background class imbalance. FCOS \cite{tian2019fcos} directly predicts at every position of the feature map and builds the anchor free detection framework.

\paragraph{Two-stage Methods.} Two-stage methods implement object detection in two steps: (\emph{i}) region proposal with raw bounding boxes and (\emph{ii}) region-based classification and bounding box refinement. For the first time, RCNN \cite{girshick2014rich} applies CNN to extract features from regions proposed by the selective search. SPPNet \cite{he2015spatial} forwards the whole image through the network and introduces spatial pyramid pooling. Fast RCNN \cite{girshick2015fast} adopts RoI-pooling, which is a differentiable layer, and thus speeds up the network significantly. Faster RCNN \cite{ren2015faster} further improves the detection efficiency by replacing the selective search with the region proposal network. R-FCN \cite{dai2016r} designs position sensitive RoI-pooling to deal with the translation-variance. Mask RCNN \cite{he2017mask} employs RoI-align instead of RoI-pooling, eliminating the misalignment caused by the quantitative operations.

Although one-stage methods have higher inference speeds, they are generally less likely to produce state-of-the-art results than two-stage methods. Since the acne detection task is insensitive to inference speed, our research mainly focuses on two-stage methods.

\subsection{Detection Head}
As the prediction module of the object detection framework, the detection head has drawn much attention. By constructing a sequence of detection heads trained with increasing IoU thresholds, Cascade RCNN \cite{cai2018cascade} achieves better object matching quality stage by stage. IoU-Net \cite{jiang2018acquisition} applies an extra branch to predict IoUs between detected bounding boxes and their corresponding ground truths. Similar to IoU-Net, Mask Scoring RCNN \cite{huang2019mask} presents mask IoU scores for each segmentation mask. Double-Head \cite{wu2020rethinking} employs two standalone branches to generate the refined bounding box and classification results. D2Det \cite{cao2020d2det} adopts a dense detection head that predicts multiple box offsets for a proposal. EFLDet \cite{liao2021efldet} proposes a trident head to integrate diverse information from the feature maps.

However, the existing methods all neglect two significant defects of the proposed detection heads: (\emph{i}) as two separate subtasks of bounding box refinement, offset and scaling are coupled in the same branch and (\emph{ii}) the potential sequential relevance of offset and scaling is not taken into account. To settle the above defects of the mainstream detection heads and promote acne detection, we implement exhaustive studies and conduct experiments on multiple datasets.

\section{Methodology}
To tackle the issues of ambiguous boundaries and arbitrary dimensions of the acne lesions, we conduct a thorough analysis on the detection heads of mainstream two-stage detectors and further develop an optimized detection head paradigm. In this section, we first elaborate the incompatibility between the offset and scaling tasks, and introduce the task-decouple mechanism. Next, we analyze the potential sequential relevance of offset and scaling, and present the task-sequence mechanism. Finally, Decoupled Sequential Detection Head (DSDH) is proposed by combining the mechanisms above, and the overall architecture is illustrated in Figure~\ref{fig_111_555}.

\subsection{Task-Decouple Mechanism}
In the two-stage detection framework, the first stage is a region proposal network (RPN), which generates raw proposal bounding boxes $B_p=\left(x_p, y_p, w_p, h_p\right)$. In order to fit the corresponding ground truth bounding boxes $B_g=\left(x_g, y_g, w_g, h_g\right)$ and the classifications $C_g$, a region-based network known as the detection head is employed in the second stage. The difference $\Delta=\left(\delta_x, \delta_y, \delta_w, \delta_h\right)$ between $B_p$ and $B_g$ is defined as
\begin{equation}
	\begin{array}{ll}
		\delta_x=\lambda_x \left(x_g-x_p\right) / w_p, & \delta_y=\lambda_y \left(y_g-y_p\right) / h_p, \\
		\delta_w=\lambda_w \log \left(w_g / w_p\right), & \delta_h=\lambda_h \log \left(h_g / h_p\right), \\
	\end{array}
	\label{eq_111}
\end{equation}
where $x_p$, $y_p$, $w_p$ and $h_p$ represent the abscissa, ordinate, width and height of $B_p$ respectively (likewise for $B_g$), and $\lambda_x$, $\lambda_y$, $\lambda_w$ and $\lambda_h$ represent the corresponding standard deviations, which are typically used as scalars 10, 10, 5 and 5 in previous works \cite{huang2017speed}.

Adopting a detection head with only a single network branch is a simple way to fit the difference $\Delta$. In this framework, a pooling operator $P$ is employed to extract proposal features $F_p$ from backbone feature maps $F$ based on proposal bounding boxes $B_p$. Then, a single branch $H$ is deployed to generate refined bounding boxes $B$ and classifications $C$ simultaneously, during which a post-processor $T$ transforms the original bounding boxes into the target bounding boxes. This framework is utilized by mainstream methods such as Faster RCNN \cite{ren2015faster} and Mask RCNN \cite{he2017mask} and summarized as Single Detection Head, whose pipeline is depicted in Figure~\ref{fig_111_111} and formulated as
\begin{equation}
	\begin{array}{lll}
		\resizebox{.87\linewidth}{!}{$
			\displaystyle
			F_p = P\left(F,B_p\right), C = H\left(F_p\right), B = T\left(B_p,H\left(F_p\right)\right).
		$} \\
	\end{array}
	\label{eq_222}
\end{equation}

Driven by the weak performance of Single Detection Head, some researchers focus on the sibling tasks, i.e., bounding box refinement and classification. They find that the essential barrier lies in the incompatibility between these two tasks. As two separate tasks, bounding box refinement and classification aim to fit two distinct inherent attributes of objects, but are tangled in the same branch. Wu \emph{et~al}.~\shortcite{wu2020rethinking} and Song \emph{et~al}.~\shortcite{song2020revisiting} introduce two standalone branches $H_1$ and $H_2$ to disentangle the sibling tasks, which effectively alleviates the incompatibility between them. This framework is summarized as Double Detection Head, whose pipeline is depicted in Figure~\ref{fig_111_222} and formulated as
\begin{equation}
	\begin{array}{lll}
		\resizebox{.89\linewidth}{!}{$
			\displaystyle
			F_p = P\left(F,B_p\right), C = H_1\left(F_p\right), B = T\left(B_p,H_2\left(F_p\right)\right).
			$} \! \! \! \\
	\end{array}
	\label{eq_333}
\end{equation}

\begin{figure}[!t]
	\centering
	\subfigure[Situations with the same scaling and different offset]{
		\includegraphics[width=0.4\columnwidth]{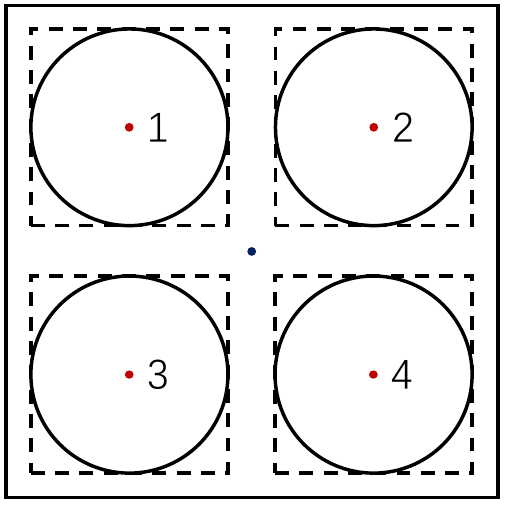}
		\label{fig_222_111}
	}
	\subfigure[Situations with the same offset and different scaling]{
		\includegraphics[width=0.4\columnwidth]{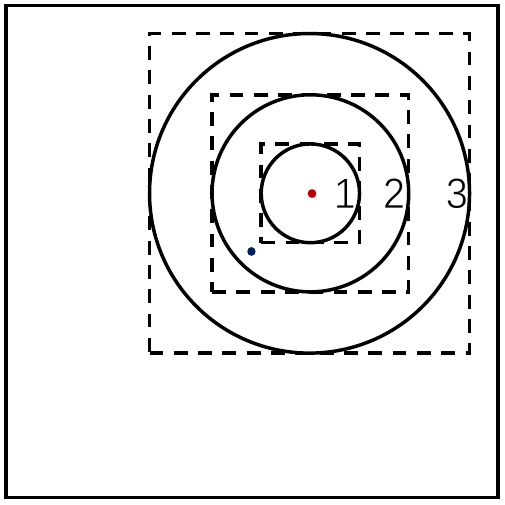}
		\label{fig_222_222}
	}
	\caption{Situations with various offset and scaling. The outer solid rectangles represent the proposal bounding boxes. The solid circles represent the simplified acne samples. The dotted rectangles represent the ground truth bounding boxes corresponding to the acne samples. The blue and red dots represent the center points of the proposals and ground truths, respectively. The numbers indicate the situations the acne samples may exhibit in the proposals.}
	\label{fig_222}
\end{figure}

Despite the merits revealed in Double Detection Head, the incompatibility between the subtasks of bounding box refinement is still neglected. We propose the \textbf{task-decouple mechanism} to settle this defect. As four separate components of the difference $\Delta$, \ $\delta_x$, $\delta_y$, $\delta_w$ and $\delta_h$ can be naturally categorized into the offset component $O=\left(\delta_x, \delta_y\right)$ and the scaling component $S=\left(\delta_w, \delta_h\right)$, where $\Delta=\left(O, S\right)$.~Through this decomposition, we divide the bounding box refinement task into the offset and the scaling subtasks. Inspired by the double-head approaches, we evaluate their incompatibility. Figure~\ref{fig_222} illustrates that whatever the situation is, the difference between the proposals and the ground truths can always be fitted by two uncorrelated subtasks, i.e., offset and scaling, representing the variations of location and size respectively. Therefore, they should not be coupled in the same branch. To further decouple offset and scaling, we deploy two standalone branches $H_2$ and $H_3$ parallelly, with the classification branch $H_1$ remaining independent. The proposed framework is denoted as Decoupled Detection Head, whose pipeline is depicted in Figure~\ref{fig_111_333} and formulated as (the classification branch $H_1$ is omitted in the formula as it is not the primary research focus, the same hereafter)
\begin{equation}
	\begin{array}{ll}
		F_p = P\left(F,B_p\right), & O = H_2\left(F_p\right), \\
		S = H_3\left(F_p\right), & B = T\left(B_p,\left(O,S\right)\right). \\
	\end{array}
	\label{eq_444}
\end{equation}

The task-decouple mechanism effectively alleviates the incompatibility between the offset task and the scaling task, and thus improves the capability of the detection head to predict the location and size of the acne lesions.

\subsection{Task-Sequence Mechanism}
Another significant defect of the mainstream detection heads is that the subtasks of bounding box refinement are executed parallelly, without their potential sequential relevance taken into account. We propose the \textbf{task-sequence mechanism} to tackle this issue. Based on the conditional probability formula, we can conclude that
\begin{equation}
	\begin{array}{c}
		p(O, S) = p(O)p(S \mid O). \\
	\end{array}
	\label{eq_555}
\end{equation}
Following this formula, the bounding box refinement task is split into two steps: (\emph{i}) generate the offset component $O$ on the basis of proposals and (\emph{ii}) generate the scaling component $S$ under the condition of $O$. Through this formulation, we execute the offset task and the scaling task sequentially.

\begin{figure}[!t]
	\centering
	\subfigure[Before executing offset]{
		\includegraphics[width=0.47\columnwidth]{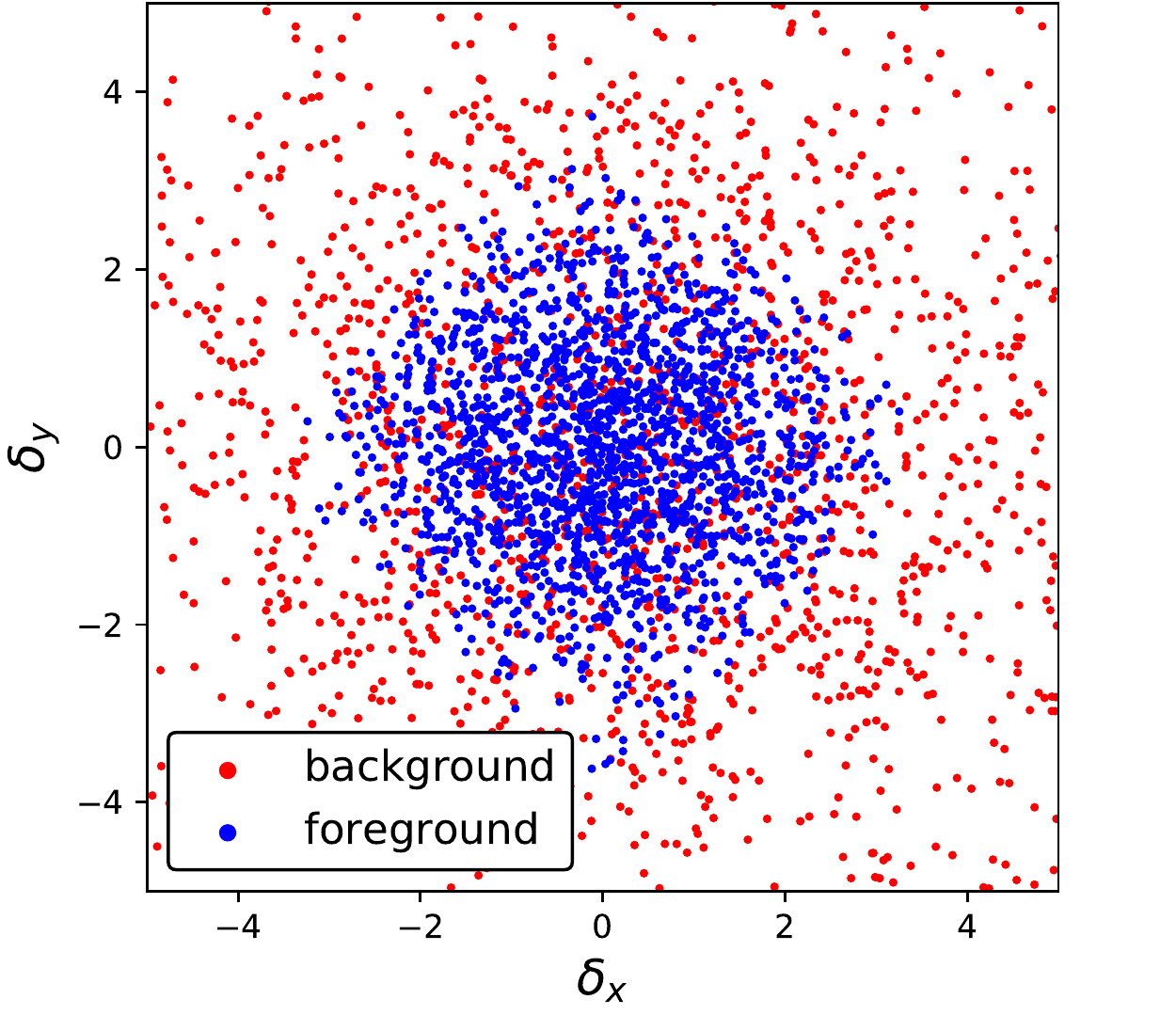}
		\label{fig_333_111}
	}
	\subfigure[After executing offset]{
		\includegraphics[width=0.47\columnwidth]{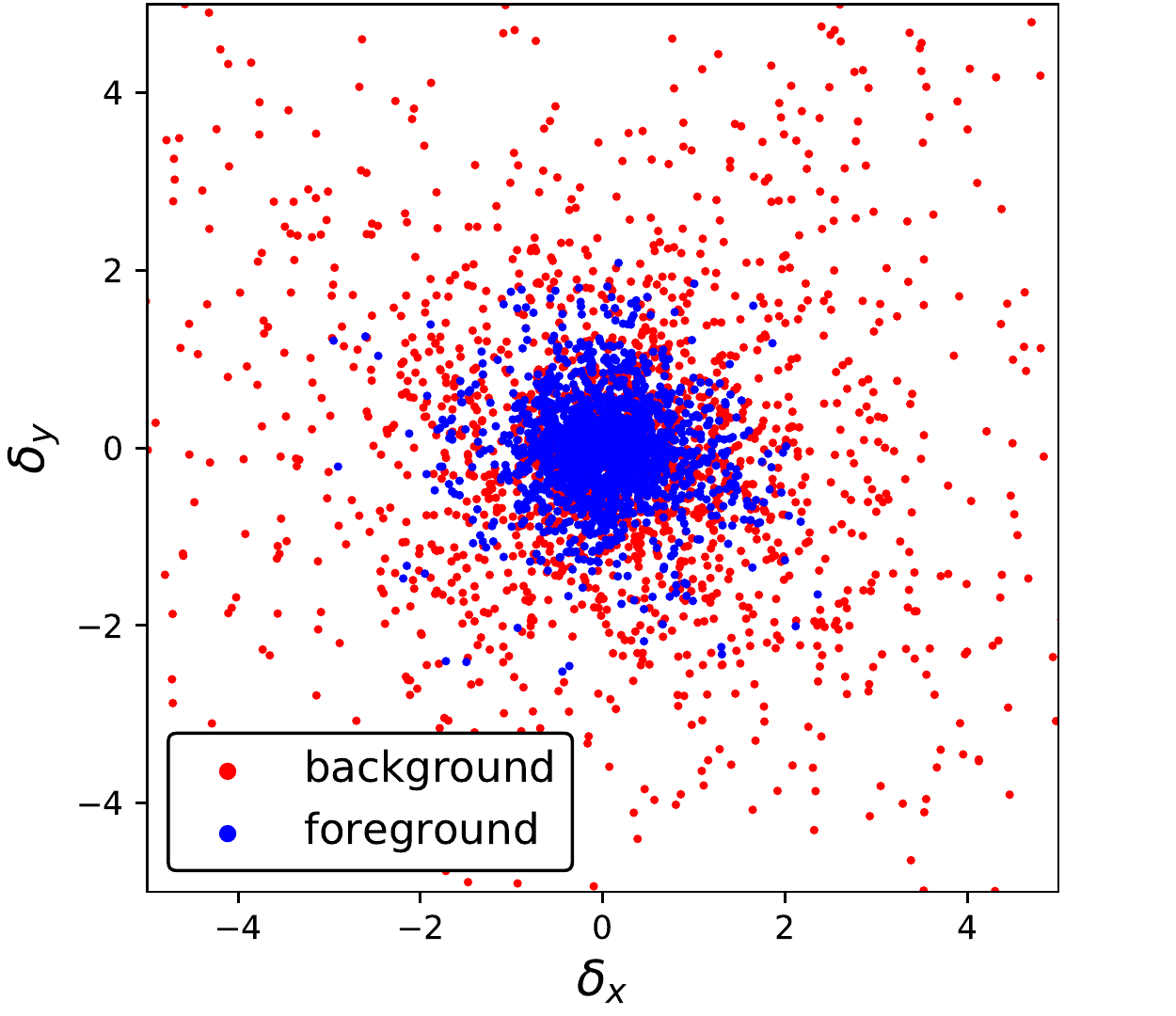}
		\label{fig_333_222}
	}
	\caption{Distributions of the components $\delta_x$ and $\delta_y$ before and after we execute the offset task. The background and foreground samples are divided by IoU=0.5 in the training phase.}
	\label{fig_333}
\end{figure}

In fact, with the help of the offset task, we can facilitate the scaling task. As shown in Figure~\ref{fig_333}, the absolute values of the components $\delta_x$ and $\delta_y$ decrease sharply after we execute the offset task, which means that the proposals get closer to the ground truths. Therefore, the scaling task is facilitated from two aspects. On the one hand, the ground truths area which appears in the proposals is increased, resulting in richer features. Intuitively, the size of an object with its entire portion in the field of view is more predictable than one with only a small portion. On the other hand, with the distributions of the ground truths converging towards the center of proposals, the requirements for the generalization capability of detection heads are also reduced.

In our sequential framework, we first execute the offset task to generate the offset component $O$, based on which the post-processor $T$ refines the proposal bounding boxes $B_p$ to $B_o$ in its location. Next, $F_o$ are extracted as the corresponding features of $B_o$. At last, we execute the scaling task to generate the scaling component $S$, based on which $T$ refines $B_o$ to the refined bounding boxes $B$ in its size. Note that all tasks share the same branch $H$ in this formulation. The proposed framework is denoted as Sequential Detection Head, whose pipeline is depicted in Figure~\ref{fig_111_444} and formulated as
\begin{equation}
	\begin{array}{lll}
		\resizebox{.80\linewidth}{!}{$
			\displaystyle
			F_p = P\left(F,B_p\right), O = H\left(F_p\right), B_o = T\left(B_p,O\right),
		$} \\
		\resizebox{.80\linewidth}{!}{$
			\displaystyle
			F_o = P\left(F,B_o\right), S = H\left(F_o\right), B = T\left(B_o,S\right).
		$} \\
	\end{array}
	\label{eq_666}
\end{equation}

The task-sequence mechanism facilitates the scaling task, enabling the detection head to gain a more comprehensive insight into the dimensions of the acne lesions.

\begin{table*}[!t]
	\setlength{\tabcolsep}{3mm}
	\centering

		\begin{tabular}{llrrrr}
			\toprule
			& Detector & AP & $AP_S$ & $AP_M$ & $AP_L$ \\
			\midrule
			\multirow{2}{*}{One-stage methods} & FCOS \cite{tian2019fcos} & 40.2 & 39.1 & 44.5 & 22.9 \\
			& RetinaNet \cite{lin2017focal} & 41.1 & 40.1 & 46.1 & 24.0 \\
			\midrule
			\multirow{6}{*}{Two-stage methods} & Cascade RCNN \cite{cai2018cascade} & 42.4 & 41.9 & 41.8 & 29.9 \\
			& Faster RCNN w/o FPN \cite{ren2015faster} & 34.4 & 29.8 & 38.6 & 29.4 \\
			& Faster RCNN \cite{ren2015faster} & 41.9 & 40.4 & 43.7 & 29.1 \\
			& Mask RCNN \cite{he2017mask} & 43.3 & 42.7 & 43.1 & 30.9 \\
			& Double-Head \cite{wu2020rethinking} & 43.8 & 44.1 & 44.3 & 29.2 \\
			& Double-FC \cite{wu2020rethinking} & 44.0 & 43.7 & 45.3 & 30.1 \\
			\midrule
			\multirow{3}{*}{\bf{Our methods}} & Cascade RCNN + DSDH & 43.8 & 42.6 & 43.9 & 31.5 \\
			& Faster RCNN + DSDH & 44.4 & 43.7 & 45.8 & 29.0 \\
			& Mask RCNN + DSDH & \bf{45.6} & \bf{44.3} & \bf{46.9} & \bf{34.1} \\
			\bottomrule
		\end{tabular}
	\caption{Comparisons with various state-of-the-art methods on \emph{ACNE-DET}. All models utilize ResNet-50 backbone.}
	\label{tab_111}
\end{table*}

\begin{table}[!t]
	\setlength{\tabcolsep}{3mm}
	\centering
	\begin{tabular}{llr}
		\toprule
		Detector & Backbone & AP \\
		\midrule
		R-FCN \cite{dai2016r} & ResNet-101 & 14.0 \\
		Rashataprucksa \emph{et~al}.~\shortcite{rashataprucksa2020acne} & ResNet-101 & 14.7 \\
		ACNet \cite{min2021acnet} & ResNet-101* & 20.5 \\
		\midrule
		Mask RCNN + DSDH\dag & ResNet-101 & \bf{23.0} \\
		\bottomrule
	\end{tabular}
	\caption{Comparisons with state-of-the-art methods on \emph{ACNE04}. *~indicates an enhanced backbone. \dag~indicates our method.}
	\label{tab_222}
\end{table}

\subsection{Decoupled Sequential Detection Head}
Utilizing the task-decouple and task-sequence mechanisms simultaneously, we propose a novel Decoupled Sequential Detection Head (DSDH). In this framework, the offset task and the scaling task are not only decoupled by two standalone branches, but also executed sequentially. Meanwhile, the classification branch remains independent. The pipeline of DSDH is depicted in Figure~\ref{fig_111_555} and formulated as
\begin{equation}
	\begin{array}{lll}
		\resizebox{.82\linewidth}{!}{$
			\displaystyle
			F_p = P\left(F,B_p\right), O = H_2\left(F_p\right), B_o = T\left(B_p,O\right),
			$} \\
		\resizebox{.82\linewidth}{!}{$
			\displaystyle
			F_o = P\left(F,B_o\right), S = H_3\left(F_o\right), B = T\left(B_o,S\right).
			$} \\
	\end{array}
	\label{eq_777}
\end{equation}

Since all the modules described above are differentiable, DSDH is trained in an end-to-end manner and optimized by the following multi-task loss
\begin{equation}
	\begin{array}{c}
		\mathcal{L} = \mathcal{L}_{cls} + \alpha \mathcal{L}_{off} + \beta \mathcal{L}_{sca}, \\
	\end{array}
	\label{eq_888}
\end{equation}
where $\mathcal{L}_{off}$ and $\mathcal{L}_{sca}$ represent the losses for the offset task and the scaling task respectively, and $\mathcal{L}_{cls}$ for the loss of the classification task. The coefficients $\alpha$ and $\beta$ are employed to balance the offset task and the scaling task. We set $\alpha=1$ and $\beta=1$ by default.

In addition, for the sake of inspecting whether DSDH can be further enhanced by the cascade architecture that is commonly used in general-purpose detection frameworks, we refer to the paradigm presented in \cite{cai2018cascade} and construct the cascade version of DSDH, whose pipeline is depicted in Figure~\ref{fig_111_666}.

DSDH aims to tackle the issues of ambiguous boundaries and arbitrary dimensions of the acne lesions through two simple but effective improvements, and it can be easily adopted by mainstream two-stage detectors.

\section{Experiments}
In this section, we first detail the datasets. Then, we describe the implementation details and evaluation metrics. Finally, several experiments are presented.

\subsection{Datasets}

\paragraph{The ACNE-DET Dataset.} Due to the absence of high-quality acne detection datasets, our work begins with the construction of a new dataset named \emph{ACNE-DET}, which has the following merits: Firstly, through VISIA, a professional facial skin image acquisition equipment, we collect a batch of high-quality skin images characterized by standardization, high-resolution and no redundant information. Second, these skin images have undergone multiple rounds of annotation and modification by six dermatologists on our specialized acne annotation system, so they have superior annotation accuracy. Finally, to exhaustively analyze the skin lesions, we divide them into ten fine-grained categories, including \emph{Close Comedo (Comedo-C)}, \emph{Open Comedo (Comedo-O)}, \emph{Papule}, \emph{Pustule}, \emph{Nodule}, \emph{Atrophic Scar (Scar-A)}, \emph{Hypertrophic Scar (Scar-H)}, \emph{Melasma}, \emph{Nevus} and \emph{Other}. They comprise both common acne and non-acne skin lesion categories. Note that the skin lesions without a certain category are labeled as \emph{Other}. \emph{ACNE-DET} contains 276 facial skin images with 31,777 labeled lesion instances. An example of the facial skin images from \emph{ACNE-DET} is shown in Figure~\ref{fig_444}. We randomly divede \emph{ACNE-DET} into a testing set and a training set which contains 241 labeled images with 28,260 skin lesions of various forms and sizes. Since the same patient may be examined multiple times, resulting in correlated data, we guarantee the data of the same patient to appear in only one dataset.

\paragraph{The ACNE04 Dataset.} In order to verify the effectiveness of DSDH, we employ \emph{ACNE04} \cite{wu2019joint} as a public benchmark of acne detection. \emph{ACNE04} contains 1,457 skin images collected by digital cameras with 18,983 bounding box annotations of a single lesion category. Following \cite{min2021acnet}, we randomly split the dataset into 80\% training set and 20\% testing set, and further crop the images into $640\times640$ sub-images.

\begin{figure*}[!t]
	\centering
	\includegraphics[width=0.98\textwidth]{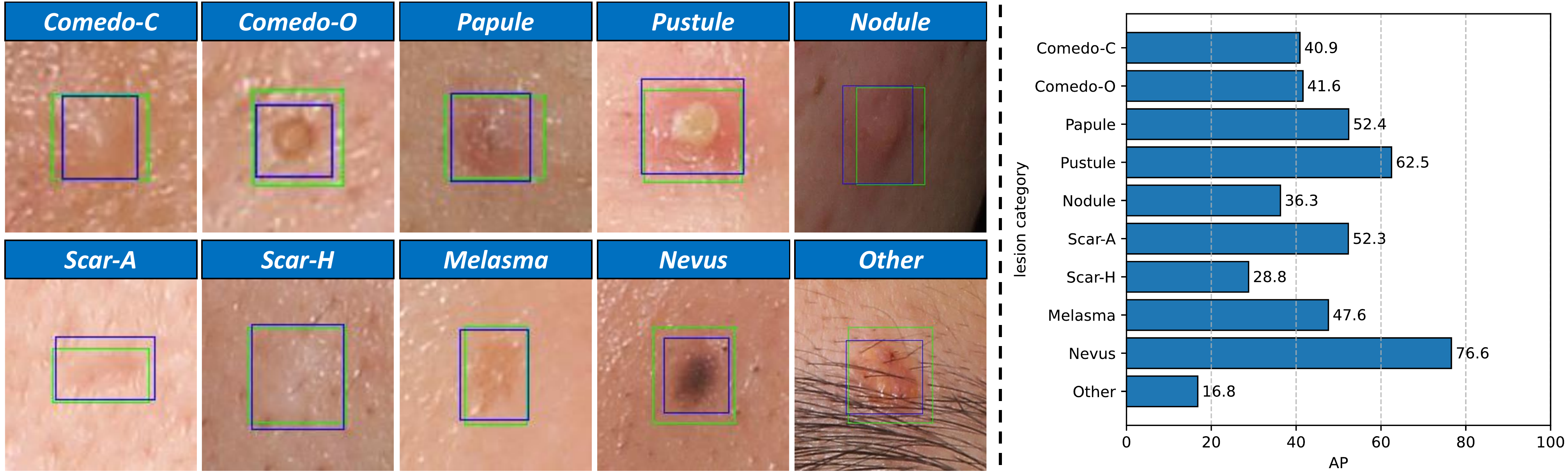}
	\caption{Visualization results by lesion category on \emph{ACNE-DET}. Left: visualization of the qualitative results. Examples of the skin lesions are cropped from the original images by category and magnified for better clarity. The bounding boxes of the ground truth and our method are marked by the green and blue rectangles, respectively. Right: visualization of the quantitative results.}
	\label{fig_555}
\end{figure*}

\begin{table}[!t]
	\centering
	\begin{tabular}{ccrrrr}
		\toprule
		Decouple & Sequence & AP & $AP_S$ & $AP_M$ & $AP_L$ \\
		\midrule
		&  & 43.3 & 42.7 & 43.1 & 30.9 \\
		& \ding{52} & 44.7 & 43.2 & 44.2 & 32.9 \\
		\ding{52} &  & 44.8 & \bf{44.5} & 44.5 & 32.9 \\
		\ding{52} & \ding{52} & \bf{45.6} & 44.3 & \bf{46.9} & \bf{34.1} \\
		\bottomrule
	\end{tabular}
	\caption{Ablation study of the model design on \emph{ACNE-DET}. The base model is “Mask RCNN + DSDH” with ResNet-50 backbone.}
	\label{tab_333}
\end{table}

\subsection{Implementation Details}
Our work is conducted based on Detectron2 \cite{wu2019detectron2} and trained on an NVIDIA 3090 GPU. The pre-trained parameters from COCO \cite{lin2014microsoft} are employed to initialize all models. The horizontal flip is used in the data augmentation, and a learning rate of $2e-3$ with a monmentum of 0.9 is adopted in the 13 epochs training. The ratio of background and foreground samples fed into the detection head is 3. NMS threshold is set to IoU=0.5 in the testing phase, and bounding boxes with classification scores greater than 0.01 are output as the detection results.

\subsection{Evaluation Metrics}
As the primary metric of object detection, mean Average Precision (AP) is commonly used to measure the performance across all categories \cite{wang2022double,xiong2022scmt}. We utilize PASCAL-style AP as the main evaluation metric of acne detection. In addition, we evaluate the performance on acne lesions in various sizes through $AP_S$, $AP_M$ and $AP_L$, which are AP for small objects (area less than $32^2$), medium objects (area between $32^2$ and $96^2$) and large objects (area greater than $96^2$), respectively. The maximum number of the bounding boxes participating in the evaluation is 100.

\subsection{Main Results}

\begin{table}[!t]
	\setlength{\tabcolsep}{3mm}
	\centering
	\begin{tabular}{ccrrrr}
		\toprule
		Conv & FC & AP & $AP_S$ & $AP_M$ & $AP_L$ \\
		\midrule
		\multirow{2}{*}{0} & 1 & 45.0 & \bf{44.5} & \bf{48.3} & 33.1 \\
		& 2 & \bf{45.6} & 44.3 & 46.9 & \bf{34.1} \\
		\midrule
		\multirow{2}{*}{1} & 1 & 44.7 & 43.7 & 46.4 & 33.1 \\
		& 2 & 44.1 & 42.9 & 45.7 & 32.6 \\
		\midrule
		\multirow{2}{*}{2} & 1 & 44.6 & 42.8 & 47.4 & 33.9 \\
		& 2 & 44.0 & 42.5 & 45.4 & 32.7 \\
		\bottomrule
	\end{tabular}
	\caption{Ablation study of the branch network on \emph{ACNE-DET}. The base model is “Mask RCNN + DSDH” with ResNet-50 backbone.}
	\label{tab_444}
\end{table}

\paragraph{Comparisons with State-of-the-art Methods.} As the representative works of two-stage detectors, Faster RCNN \cite{ren2015faster} and Cascade RCNN \cite{cai2018cascade} are employed in the experiment on \emph{ACNE-DET}. Mask RCNN \cite{he2017mask} is also included as an enhanced version of Faster RCNN. Then, we replace the original detection heads of the above detectors with DSDH to verify its effectiveness. Another state-of-the-art detector, namely Double-Head, and its variation Double-FC \cite{wu2020rethinking} are also utilized. Besides, FCOS \cite{tian2019fcos} and RetinaNet \cite{lin2017focal} are chosen as the representative works of the one-stage detectors. Note that we try our best to implement the compared methods with the same hyper-parameters and settings. As shown in Table~\ref{tab_111}, our method achieves state-of-the-art results compared with both one-stage and two-stage~methods, utilizing the same ResNet-50 backbone. Specifically, Mask RCNN based DSDH outperforms all other methods, and improves AP by 2.3\%, $AP_S$, $AP_M$ and $AP_L$ by 1.6\%, 3.8\% and 3.2\% respectively compared with the original Mask RCNN. It~should be noted that this is actually a significant progress. In fact, even dermatologists sometimes fail to label consistent ground truths due to the ambiguous boundaries of the acne lesions, thus severely impeding the performance gains. Furthermore, Faster RCNN based DSDH and Cascade RCNN based DSDH surpass the original versions by 2.5\% and 1.4\% respectively, demonstrating the generalization capability of our method. We believe DSDH effectively alleviates the incompatibility between the offset and scaling tasks, and further facilitates the scaling task in acne detection. Notably,~two-stage detectors achieve better results than one-stage detectors, especially on $AP_L$. This may attribute to the universality of the two-stage detectors. It is noticed that, with or without~DSDH, the cascade detectors perform worse than the vanilla ones. We argue that the increasing IoU thresholds are unsuitable for acne detection due to the ambiguous boundaries.

\begin{table}[!t]
	\setlength{\tabcolsep}{3mm}
	\centering
	\begin{tabular}{lrrrr}
		\toprule
		Backbone & AP & $AP_S$ & $AP_M$ & $AP_L$ \\
		\midrule
		ResNet-50 & \bf{45.6} & \bf{44.3} & \bf{46.9} & \bf{34.1} \\
		ResNet-101 & 45.2 & 43.9 & 46.2 & 32.5 \\
		ResNeXt-101 & 42.8 & 44.2 & 45.0 & 26.4 \\
		\bottomrule
	\end{tabular}
	\caption{Ablation study of the backbone scale on \emph{ACNE-DET}. The base model is “Mask RCNN + DSDH.”}
	\label{tab_555}
\end{table}

\paragraph{Visualization Results by Lesion Category.} The results of our Mask RCNN based DSDH are visualized in Figure~\ref{fig_555} to illustrate the detection performance for each lesion category. We can see that our method works well for most categories, and even fits the skin lesion areas better than the ground truths labeled by the dermatologists in some cases, which exhibits the high performance of the proposed DSDH on acne detection. Note that AP of \emph{Other} is relatively low since it consists of various lesion categories.

\paragraph{Comparison Results on ACNE04.} To further demonstrate the effectiveness of our method, we compare Mask RCNN based DSDH with the detection methods that have published results on \emph{ACNE04}. As shown in Table~\ref{tab_222}, our method outperforms all the prior state-of-the-art methods by large margins, without bells and whistles.

\subsection{Ablation Studies}
We perform groups of ablation studies on \emph{ACNE-DET} to analyze the optimal design of our method. Mask RCNN based DSDH is chosen as the base model in the following experiments for its best performance.

\begin{table}[!t]
	\setlength{\tabcolsep}{1.5mm}
	\centering
	\begin{tabular}{lrrrr}
		\toprule
		Decouple Manner & AP & $AP_s$ & $AP_m$ & $AP_l$ \\
		\midrule
		No decouple \cite{he2017mask} & 43.3 & 42.7 & 43.1 & 30.9 \\
		Double-FC \cite{wu2020rethinking} & 44.0 & 43.7 & 45.3 & 30.1 \\
		\midrule
		Offset plus scaling\dag & \bf{44.8} & \bf{44.5} & 44.5 & \bf{32.9} \\
		Horizontal plus vertical\dag & 44.3 & 44.2 & \bf{46.0} & 32.5 \\
		Fully decoupled\dag & 44.3 & 44.0 & 45.1 & 31.0 \\
		\bottomrule
	\end{tabular}
	\caption{Evaluations of various decouple manner on \emph{ACNE-DET}. “No decouple” represents Mask RCNN. \dag~indicates our variations.}
	\label{tab_666}
\end{table}

\paragraph{Model Design.} We omit different components in DSDH, including the task-decouple mechanism and task-sequence mechanism, to investigate their effectiveness. Results are shown in Table~\ref{tab_333}. The task-decouple mechanism gains AP by 1.5\%, while the task-sequence mechanism leads to a gain of 1.3\%. These two mechanisms both prove their ability to enhance acne detection. Furthermore, they can achieve mutual benefit when combined.

\paragraph{Branch Network.} We design the structure of the branch networks by exploring the layer numbers, on the premise that the channel numbers of the convolutional layers and the fully connected layers are fixed at 256 and 1024, respectively. From Table~\ref{tab_444}, we can learn that with the number of convolutional layers increasing, the performance of the model declines. We adopt “Conv-0 FC-2” in all branches for its superior performance. Note that this is the default setting of the two-stage methods shown in Table~\ref{tab_111}, except Double-Head.

\paragraph{Backbone Scale.} Table~\ref{tab_555} shows the comparisons of different backbone scales. ResNet-50, ResNet-101 and ResNeXt-101 \cite{xie2017aggregated} 32x8d are utilized as the competitors, among which ResNet-50 takes the lead on all metrics.

\subsection{Structure Variations}
To gain a more comprehensive understanding of our approach, we explore the variations of the task-decouple and the task-sequence manners. Explorations are based on Mask RCNN with ResNet-50 backbone on \emph{ACNE-DET}.

\paragraph{Variations of Task-Decouple.} We discuss the variations of task-decouple manner of the bounding box refinement task. Three variations are explained as follows:

\begin{itemize}
	\item Offset plus scaling: the bounding box refinement task is divided into the offset and the scaling subtasks, which are decoupled by two standalone branches (our choice).
	\item Horizontal plus vertical: the bounding box refinement task is divided into the horizontal and the vertical subtasks, which are decoupled by two standalone branches.
	\item Fully decoupled: the subtasks of bounding box refinement are fully decoupled by four standalone branches.
\end{itemize}
where the horizontal task and the vertical task consist of the subtasks relevant to x-axis and y-axis, respectively (i.e., decompose $\Delta$ into $\mathcal{H}=\left(\delta_x, \delta_w\right)$ and $\mathcal{V}=\left(\delta_y, \delta_h\right)$). Note that the task-sequence mechanism is not implemented, and the classification branch is independent in the above variations.

Results are shown in Table~\ref{tab_666}. “No decouple” (i.e., Mask RCNN) and Double-FC are listed as the baseline methods. We can learn all variations achieve performance gains over the baseline methods, with “Offset plus scaling” taking the lead. Results demonstrate that “Offset plus scaling,” adopted in our task-decouple mechanism, is the optimal option and alleviates the incompatibility between offset and scaling.

\begin{table}[!t]
	\centering
	\begin{tabular}{lrrrr}
		\toprule
		Sequence Manner & AP & $AP_s$ & $AP_m$ & $AP_l$ \\
		\midrule
		Parallel \cite{he2017mask} & 43.3 & 42.7 & 43.1 & 30.9 \\
		\midrule
		Offset then scaling\dag & \bf{44.7} & \bf{43.2} & 44.2 & \bf{32.9} \\
		Scaling then offset\dag & 44.2 & \bf{43.2} & 45.3 & 31.6 \\
		Horizontal then vertical\dag & 44.0 & 42.9 & \bf{46.1} & 31.7 \\
		Vertical then horizontal\dag & 44.2 & 42.9 & 45.9 & 32.3 \\
		\bottomrule
	\end{tabular}
	\caption{Evaluations of various sequence manner on \emph{ACNE-DET}. “Parallel” represents Mask RCNN. \dag~indicates our variations.}
	\label{tab_777}
\end{table}

\paragraph{Variations of Task-Sequence.} The variations of the task-sequence manner are analyzed after we split the bounding box refinement task into two steps. Four variations are explained as follows:

\begin{itemize}	
	\item Offset then scaling: execute the offset task first, then the scaling task (our choice).
	\item Scaling then offset: execute the scaling task first, then the offset task.
	\item Horizontal then vertical: execute the horizontal task first, then the vertical task.
	\item Vertical then horizontal: execute the vertical task first, then the horizontal task.
\end{itemize}
Note that the task-decouple mechanism is not implemented in the above variations, which means that all tasks share the same branch.

As shown in Table~\ref{tab_777}, all variations obtain better results compared with the baseline method “Parallel” (i.e., Mask RCNN), among which “Offset then scaling” performs best. Results verify our opinion that “Offset then scaling” effectively facilitates the scaling task, and is the best choice for our task-sequence mechanism.

\section{Conclusion and Future Research}
In this paper, we propose a novel Decoupled Sequential Detection Head (DSDH) for acne detection. DSDH tackles the issues of ambiguous boundaries and arbitrary dimensions of the acne lesions via two simple but effective improvements. Our task-decouple mechanism settles the incompatibility between the offset and scaling tasks to improve the capability of predicting the location and size of acne lesions. Utilizing the task-sequence mechanism, we execute offset and scaling sequentially to gain a more comprehensive insight into the dimensions of acne lesions. In addition, we construct a high-quality acne detection dataset named ACNE-DET to verify the effectiveness of DSDH. Experiments show our method significantly outperforms the state-of-the-art methods. 

DSDH optimizes the architectures of conventional detection heads, and can be easily adopted by mainstream two-stage detectors. In future research, it is worth exploring the capability of DSDH as a general-purpose detection head.

\bibliographystyle{0_other/named}
\bibliography{0_other/ref}

\end{document}